# Machine Learning Approaches for Binary Classification to Discover Liver Diseases using Clinical Data

*Fahad B. Mostafa*[1] **and** *Easin Hasan*[2]

1. Department of Mathematics and Statistics, Texas Tech University, Lubbock, TX-79409.
2. Department of Mathematical Sciences, The University of Texas at El Paso, El Paso, TX-79968.
`{Emails: fahad.mostafa@ttu.edu; mhasan8@miners.utep.edu}`

## ABSTRACT

For a medical diagnosis, health professionals use different kinds of pathological ways to make a decision for medical reports in terms of patients' medical condition. In the modern era, because of the advantage of computers and technologies, one can collect data and visualize many hidden outcomes from them. Statistical machine learning algorithms based on specific problems can assist one to make decisions. Machine learning data driven algorithms can be used to validate existing methods and help researchers to suggest potential new decisions. In this paper, Multiple Imputation by Chained Equations was applied to deal with missing data, and Principal Component Analysis to reduce the dimensionality. To reveal significant findings, data visualizations were implemented. We presented and compared many binary classifier machine learning algorithms (Artificial Neural Network, Random Forest, Support Vector Machine) which were used to classify blood donors and non-blood donors with hepatitis, fibrosis and cirrhosis diseases. From the data published in UCI-MLR, all mentioned techniques were applied to find one better method to classify blood donors and non-blood donors (hepatitis, fibrosis, and cirrhosis) that can help health professionals in a laboratory to make better decisions. Our proposed ML-method showed a better accuracy score (e.g. 98.23% for SVM). Thus, it improved the quality of classification.

# 1. INTRODUCTION

Machine Learning (ML) algorithms are new techniques to handle many hidden problems in medical data sets. This approach can help health-care management and professionals to explore better results on many clinical applications such as medical image processing, language processing, tumor or cancer cell detections by finding appropriate features. To detect disease, lab workers need to collect different samples from patients which takes more time and costs. Medical data costs a lot of money and it is very expensive to run trials and to find patients willing to volunteer, especially for rare diseases. To address such problems, several statistical and engineering approaches (e.g. simulation modelling, classification, inference etc.) have been used by researchers and lab technicians (Beneyan., 1997, Daugury et. al., 2007, Subramaniyan et. al., 2016). The results are more data driven than model dependent. In medical reports, finding appropriate targets (Esteva et. al., 2017) from features is a challenging classification problem. Logistic regression is a widely used technique, but its performance is relatively poorer than several Machine Learning and Deep Learning methods (Couronné et al., 2018; Musa 2013; Dreiseitl et al., 2002). First of all, data visualization is necessary to understand latent knowledge about predictors, which is a part of exploratory data analysis (Seo et. al., 2004). Among many techniques, whiskers plot is one of them, which indicates variability outside the upper and lower quantiles, which are known as outliers. Another common problem in real life application of data science is missing values in the data set. To handle them, one can use several methods. However, wrong imputation of missing values can lead models toward wrong prediction. MICE is known as Multiple Imputation by Chained Equations, which technique helps to manipulate missing variables (Raghunathan et. al., 2001; Buuren et. al., 2007). It gives the assumption of missing data at a random procedure which is investigated as Missing At Random (MAR) method. It implies that the probability of missing value depends only on observed values, not the values which are not observed (Graham et. al., 2002). This procedure creates numerous predictions for each missing value with multiply imputed data taking into consideration uncertainty in the imputations and produces some accurate standard errors. MICE algorithm (Chowdhury et al., 2017) is a nice performer between many data imputation methods. A heatmap (Wilkinson et. al., 2009) is another way to see correlation between input variables. Moreover, medical disease detections are mostly reliable on biological and biochemical features where all of them are not significant. For optimal feature section, Principal Component Analysis (PCA) is a conventional technique to reduce dimensionality in medical diagnosis (Pechenizkiy et. al., 2004; Mostafa et. al., 2019). Many researchers from different fields studied binary classification using machine learning (Jafri et. al., 2018, Esteva et. al., 2017) for detecting breast cancer, skin cancer and many other problems related to health diagnosis. One of the used methods is the Support Vector Machine (Schölkopf et. al., 1996). It was inaugurated by Boser, Guyon, and Vapnik in COLT-92 (Vapnik 1995). This helps us to divide the label by the hypersphere of a linear function in a high dimensional feature space, which is developed with a learning algorithm from mathematical optimization where the learning bias will be calculated using statistical learning. Support Vector Machine (SVM) can assist to make decisions using maximum linear classifiers with the highest range (Berges 1998, Nello et. al., 2000). Another model for classification problems is the Artificial Neural Network (ANN). The concept of ANN is similar to neurons in the human body. Machine Learning for breast cancer detection with ANN was studied by (Jafari et al. 2018). Another method which outperforms decision tree algorithms is Random Forests (RF) (Ho. 1995). This ML procedure was studied by many researchers for binary classification of cancer data, X-ray image data for pattern recognitions (ko. 2011, De Oliveira et. al., 2018). Pianykh (Pianykh et al., 2018) studied healthcare operation management using Machine Learning. In their paper they mentioned several future directions. In this paper several research questions should be considered; firstly, how to handle missing data and unnecessary predictors? Secondly, does ML approach perform better for medical diagnosis such as liver disease? To answer these problems, ML methods had been introduced to optimize liver disease from some medical reports, and based on this approach, blood donors were identified. We started with data visualization techniques (Meyer et. al., 1991) to plot the summary of the input variables. Before doing that, we looked at the data to find missing values. Using MICE (Friendly, 2008), we imputed missing data. For every predictor, the split of all possible values was considered. We had several predictors and one response (which we tried to predict). Response variable was considered as the person of being a blood donor or not. Using real life data set [1], our study showed how machine learning algorithms can improvise medical professionals' ability to manage medical reports in a more advanced and effective way.

**DATA DETAILS**

Data was collected from the University of California Irvine Machine Learning Repository (UCI-MLR) [1]. Data set is included with laboratory reports of blood donors, and non-blood donors with Hepatitis C patients and demographic values like age. The response variable for classification was Category type (blood donors vs. Hepatitis C (including its progress ('just' Hepatitis C, Fibrosis, Cirrhosis). It was contained 14 attributes such as Albumin (ALB), Alkaline Phosphatase (ALP), Bilirubin (BIL), Choline Esterase (CHE), Gamma Glutamyl-Transferase (GGT), Aspartate Amino-Transferase (AST), Alanine Amino-Transferase (ALT), Creatinine (CREA), Protein (PROT), and Cholesterol (CHOL). On the other hand, sex and diagnosis were categorical, and age was nominal data. In the data, there are 615 observations are collected.

## 2. METHODS

Statistical machine learning methods are implemented to diagnose liver disease from medical reports. A flow chart has been presented below to show the overview of the research.

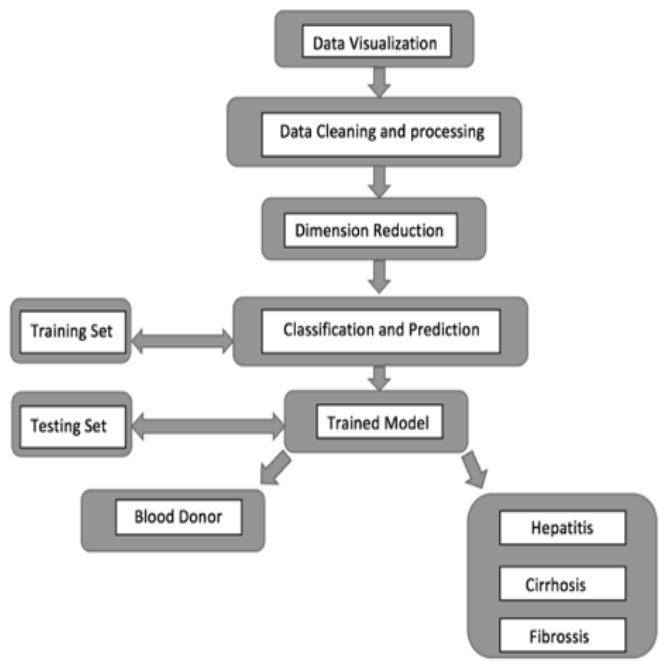

**Figure 1: Flow Chart of Research Method.**

**DATA VISUALIZATIONS and TARGET LABELING:**

Missing data is quite a common scenario in the application of data science. In this study, we dig into data by using different plots. Our main aim was to detect groups of people who donated blood and people who did not because of disease. The target variable has been modified into a binary category so that blood donors label as category "0" and it falls into either "0=Blood Donor" or "0s=suspect Blood Donor" and 1 if it falls into any other category with liver disease. The following method was used to fill out missing items for each predictor in our multivariate data.

## 2.1 VARIABLE SELECTION and DIMENSION REDUCTION:

Let, $D \epsilon R^{N \times K}$ is the data set where $N$ is the dimension predictors and $K = \{0,1\}$ is the number of responses. In this study, two responses have been identified. We need to determine missing values so that the data set remains in balance.

### 2.1.1 Multiple Imputation by Chained Equations for Missing Data:

Multiple Imputation was used by the Chained Equations method to create the missing data as in (Buuren et. al. 2010). For multivariate missing data we have considered a R-package known as 'mice' which creates multiple imputations. This function auto detects certain variables with missing values. It basically uses predictive mean matching (PMC) which is a semi parametric imputation. It is very close to regression except missing items are randomly filled by regression prediction. The chained equation procedure has been divided down into five steps (Azur et. al. 2011)

**MICE steps**

Step 1: Start with imputing the mean. Mean imputations are considered as "position holders."
Step 2: The "position holder" presents imputations for one variable ("Var") that are impeded to the missing items.
Step 3: "Var" is the response variable where the other variables are predictor variables in the linear regression model (under the same assumption).
Step 4: The missing values for "Var" are then replaced with imputed values from the regression model.
Step 5: Repeat steps 2–4 and produce missing data. One iteration is needed for each variable, and finally missing values. 10 such cycles have been performed (Raghunathan et. al., 2002).

### 2.1.2 Principal Component Analysis:

For dimension reduction, PCA is a very well recognized feature reduction technique. Let us consider data points $x_1, x_2, \ldots, x_k \in R^K$. It is convenient to consider that the points are centered at the $\sum_i^k x_i = 0$. Then, the mean is calculated $\mu = \frac{1}{k}\sum_i^k x_i$ and $x_i = x_k - \mu$; It wants to represent these points in the lower dimensional space, suppose that $R^d$; such that $d \ll K$ must be satisfied. So, form the multivariate data, the sample covariance matrix is $S$. Where,

$$S = \frac{1}{k}\sum_i^K x_i x_i^T \qquad (2.1)$$

To find eigen decomposition, S has been decomposed as follows,

$$S = P\Lambda P^T \qquad (2.2)$$

Where the columns of $l$ are the eigenvectors of $l_1, l_2, \ldots, l_K$, the diagonal elements of $\Lambda$ are the eigenvalues. Considering the eigenvalues are sorted from large to small. Then first $d$ eigen vectors $l_1, l_2, \ldots, l_K$ has been chosen. Thus, the new representation of any $x_i$ is as follows

$$\left[l_1^T x_i, l_2^T x_i, \ldots, l_d^T x_i\right]^T \qquad (2.3)$$

According to statistical machine learning technique, the data set (collected from Hoffmann et. al. 2018) was divided into the training and testing datasets where the training set was applied to fit the parameters. In fact, it was split into training and test dataset based on v-fold cross-validation on the mis-classification error, with $v = 10$.

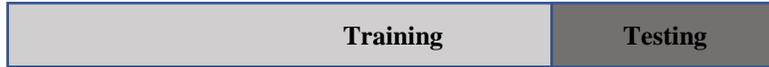

**Figure 2: ML procedure of data splitting process**

## 2.2 CLASSIFICATION MODELS:

This section includes the binary classification methods for liver disease data.

### 2.2.1 Support Vector Classifier:

SVM can be used for finding pattern recognition by classification and regression. In the class recognition problem, we start working with training data of the form

$$(x_1, y_1), \ldots, (x_k, y_k) \in R^k \times \{+1, -1\} \qquad (2.4)$$

Here, k-dimensional patterns vectors are $x_i$ and their labels are $y_i$. The label with the value of +1 states that the vector is classified to class +1 and a label of −1 states that the vector is part of class −1.

Here, we have two levels, +1 for blood donors and -1 for non-blood donors. Now we try to optimize a function $f(x) = y : R^n \to \{+1, -1\}$. The label with the value of +1 denotes that the vector is classified to class +1 and a label of −1 denotes that the vector is part of class −1.

$$\langle w. x \rangle + bias = 0; where\ w\ \epsilon\ R^k\ and\ bias\ \epsilon\ R \quad (2.5)$$

this divides the input space into two spaces, where one space contains vectors of the class −1 and another contains class +1. We can show separations of labels using hyperplane. To find a class from the particular vector x, we apply the following decision function

$$f(x) = sign(\langle w. x \rangle + bias) \quad (2.6)$$

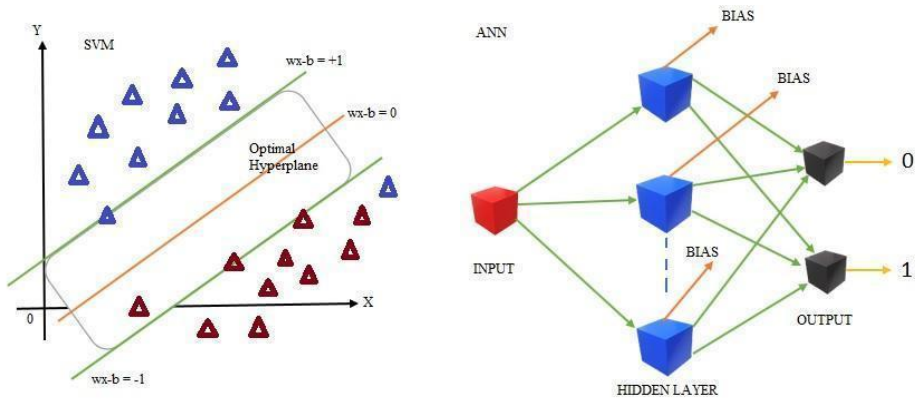

**Figure 3: SVM (left) and ANN (right)**

## *2.2.2 Artificial Neural Network Classifier:*

Now the Artificial neural network has been applied, it works similarly to the human brain's neural network. A "neuron" in a neural network is a mathematical function that collects and classifies information according to a specific architecture. A common activation function that is used is the sigmoid function:

$$f(x) = \frac{1}{1 + e^{-x}} \quad (2.7)$$

ANN helps to map the feature-target relations. It is made up with the layers of neurons where each of them works as a transformation function. The most important step is training which involves minimization of a cost function. At the end Once training is finished and validated, the application is cheap and fast. In the learning phase, the network learns by adjusting the weights to predict the correct class label of the given inputs.

$$f(x) = \sum_i^p w_i * x_i + \beta \qquad (2.8)$$

In this model, $x_1, x_2, x_3 \ldots x_p$ are input variables. $w_1, w_2, w_3, \ldots, w_p$ are weights of respective inputs. $\beta$ is the bias, which is added with the weighted inputs to form the net inputs accurately. Bias and weights are both adjustable parameters of the neuron. Equation (7) and (8) help us to determine the output with two labels. Now, for instance $f^{[1](i)}$ is the output from $i - th$ neuron of the first layer. Therefore,

$$f^{[1](i)}(x) = w^{[1]}x^{(i)} + \beta^{[1](i)} \qquad (2.9)$$

Need to pass this equation (9) through the tangent hyperbolic activation function to get

$$\alpha^{[1](i)}(x) = tanh\,(f^{[1](i)}(x))$$

So, the output layer will be

$$f^{[2](i)}(x) = w^{[2]}\alpha^{[1](i)}(x) + \beta^{[2](i)} \qquad (2.10)$$

Then finally we pass equation (9) through the sigmoid (7) activation function and we calculate our output probability as follows

$$\hat{y}^{(i)} = \alpha^{[2](i)}(x) = \sigma(f^{[2](i)}(x))$$

We use the following piecewise function to obtain predictive class from output probabilities

$$y_{pred.}^{(i)} = \begin{cases} 1, & if\ \alpha^{[2](i)}(x) > 0 \\ 0, & o.w. \end{cases} \qquad (2.11)$$

### 2.2.3 Random Forest Classifier:

Now, another method has been applied. Random forests (Breiman, 2001) is a substantial modification of bagging that builds a large collection of de-correlated trees, and then we average them. RF is very similar to boosting, and easy to train and tune. We use an average of $p$ identical and independent random variables having variance $\sigma^2$. Random forest helps us to improve the variance reduction of bagging by reducing the correlation between the trees, without increasing the variance too much. We consider, any $p$ from 1 to $P$; where we take bootstrap sample $W^*$ of size $P$ from training data. Then we grow a random-forest tree $T_p$ to the bootstrapped data. Now we repeat the following process for each terminal node of the tree, until we get the minimum node size $n_{min}$. The we find $m$ variables at random from the $p$ variables and divide the node into two daughter nodes. Finally, we find the ensemble of the trees by presenting the sequence $\{T_p\}_1^P$. Them we observe our prediction at a new point $x$. So, for classification, we consider $\widehat{C_p}(x)$ be the class prediction of the $pth$ random-forest tree. Thus,

$$\widehat{C}^P(x) = majority\ vote\{\widehat{C_p}(x)\}_1^P \qquad (2.12)$$

## 2.4 EVALUATION OF MODELS

In order to describe the accuracy of a binary classification model, we often use the measures of precision sensitivity and specificity. Accuracy is the model's ability to correctly identify the observations while the precision measures the model's ability to distinguish between positive and negative observations. The sensitivity measures how many positive classifications are found of all the available positive classifications while the specificity has the same interpretation for the negative observations. Components of the Confusion Matrix are True Positive (TN), False Positive (FP), True Negative (TN), and False Negative (FN).

| Confusion Matrix | Actual Class | | |
|---|---|---|---|
| **Predicted Class** | Model | 0 | 1 |
| | 0 | *TP* | *FN* |
| | 1 | *FP* | *TN* |

$$Accuracy = \frac{TP + TN}{TP + TN + FP + FN}$$

$$Precision = \frac{TP}{TP + FP}$$

$$Sensitivity = \frac{TP}{TP + FN}$$

$$Specificity = \frac{TN}{TN + FP}$$

$$F_1 = \frac{2 \times Precision \times Recall}{Precision + Recall}$$

In the above formula and in confusion matrix, TPs are true positives, TNs are true negatives, FPs are false positives, and FNs are false negatives. The confusion matrices indicate the percentage of correct and incorrect classification of each class which indicates exactly between what classes, the algorithms have the most difficulties for classification for trained models. TP and TN indicate the actual number of data points of positive class and negative class, respectively, that the model also labeled as True. Finally, FP indicates the number of negatives that the model classified as positives and FN represents the number of positives that the machine classified as negatives.

To visualize the performance of our model we will use the (Receiver Operating Characteristic) ROC curve. It plots the sensitivity against the 1-specificity for different cuts of points. On the other hand, Area Under the Curve (AUC) is a process to summarize the performance of a model with just one value. The magnitude is the area under the ROC curve and is a ratio between 0 and 1 where a value of 1 is a perfect classifier while a value close to 0.5 is a bad model since that is equivalent to a random classification from the train set. Moreover, in case of RF, gini index is related to ROC such that, gini index is the area between ROC curve and no-discrimination line (linear) times two. That is, the formula for gini index is,

$$G_i = 2AUC - 1.0 \qquad (2.13)$$

After selecting features using PCA, we divide the reduced data sets into two parts where 564 persons are selected training data and 51 persons are in the test data set. Supervised learning was carried out on the dataset using random forest and support vector machines. A ROC curve, with False Alarm Rate (x-axis) vs Hit Rate (y-axis) was plotted for ANN, RF and SVM. From Table 01, highest accuracy has been found for SVM. Variance importance ranking table uses mean decrease accuracy and mean decrease gini index to determine which variables are important. Then, a variable importance ranking plot was also plotted. On the x-axis is false alarm rate which is anomalous to false positive and on the y-axis is hit rate which is anomalous to true positive. Thus, ROC could be regarded as a plot of power as a function of Type I error. Mean decrease accuracy and gini index determine which variables are important as, in the plot, from top to bottom, the most important and least important variables are ranked. And, variables with large mean decrease value are important. Further, gini index measures homogeneity of variables as compared with original data. Further, gini index is related to ROC such that, gini index is the area between ROC curve and no-discrimination line (linear) times two. If $\hat{y}$ is the class level predicted by our applied models where $y$ is the correct class level, then the loss function,

$$L(\hat{y}, y) = \begin{cases} 1, & if \ \hat{y} = y \\ 0, & if \ \hat{y} \neq y \end{cases}$$

The predicted error from test data set has found from the following equations

$$Error = E[L(\hat{y}, y)]$$

$$Error_{test} = \frac{1}{n}\sum_{i=1}^{n} L(\hat{y}^i, y^i)$$

## 3. RESULTS AND DISCUSSION:

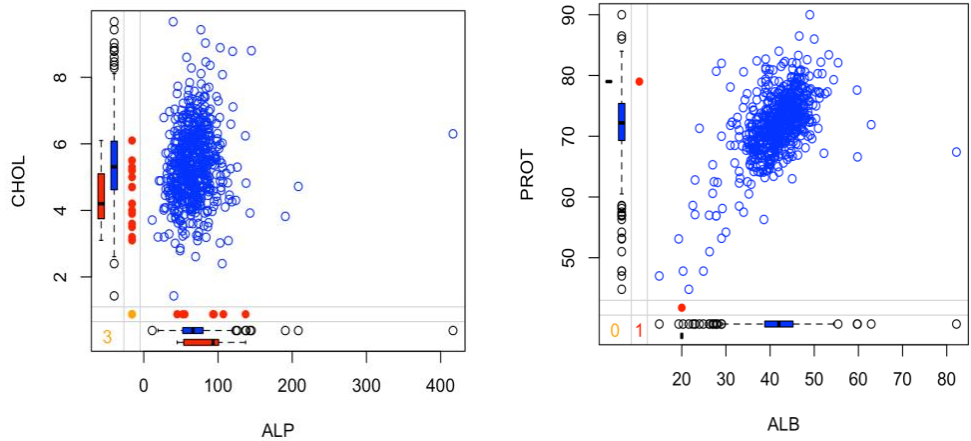

**Figure 4: Margin-plot for pattern and distribution of complete and incomplete observations in missing features**

The **Figure 4** helps us to investigate the pattern as well as the distribution of incomplete and complete observations of missing input variables. For the plot of left-side ALP and CHOL and right-side ALB and PROT, blue dots represent observations. In the left and bottom margins, blue box plots are non-missing and red box plots are the marginal distribution of these observed values. From the exploratory data analysis, ALP, ALB, CHOL and PROT have missing values.

There are several discrepancies between the range and variation of predictors with many outliers. We figured out there are some extreme outliers for the following variables: ALT, AST, CREA, and GGT. Thus, **Figure 5** indicates that some donors have high amounts of ALT, AST, CREA, and GGT in their bloods.

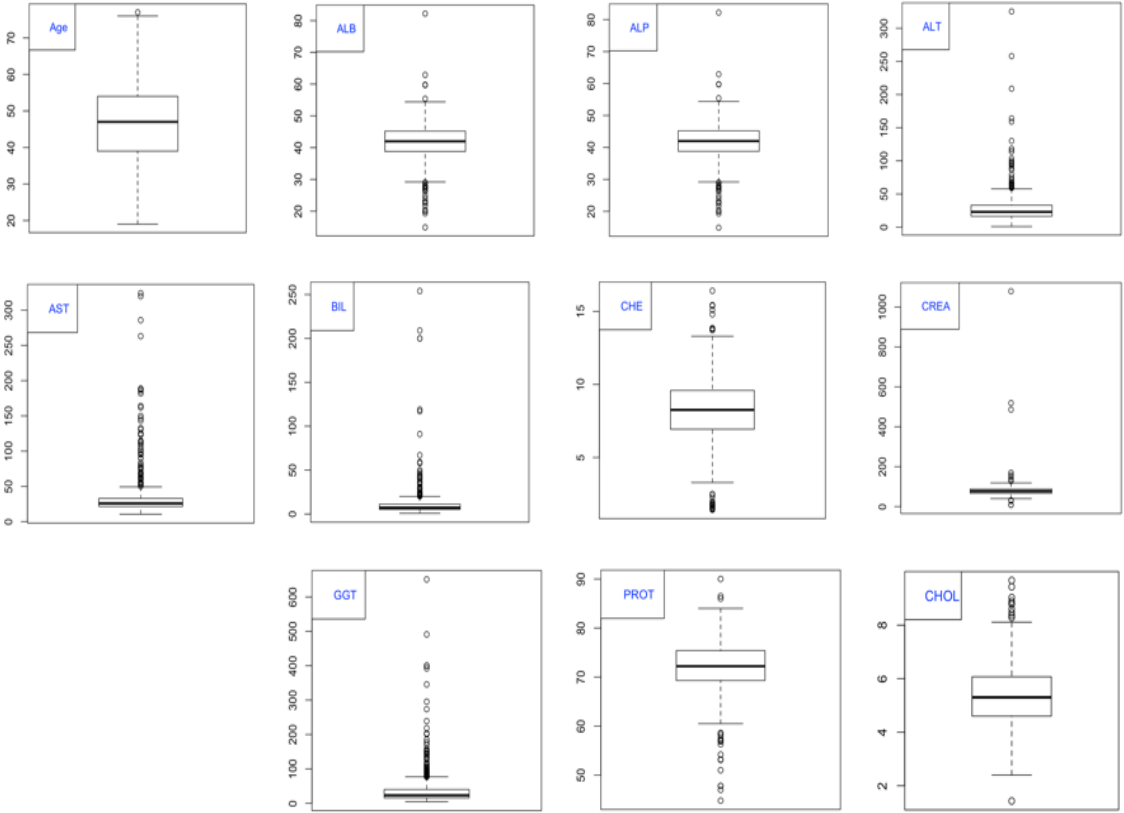

**Figure 5: Box plot for predictor variables**

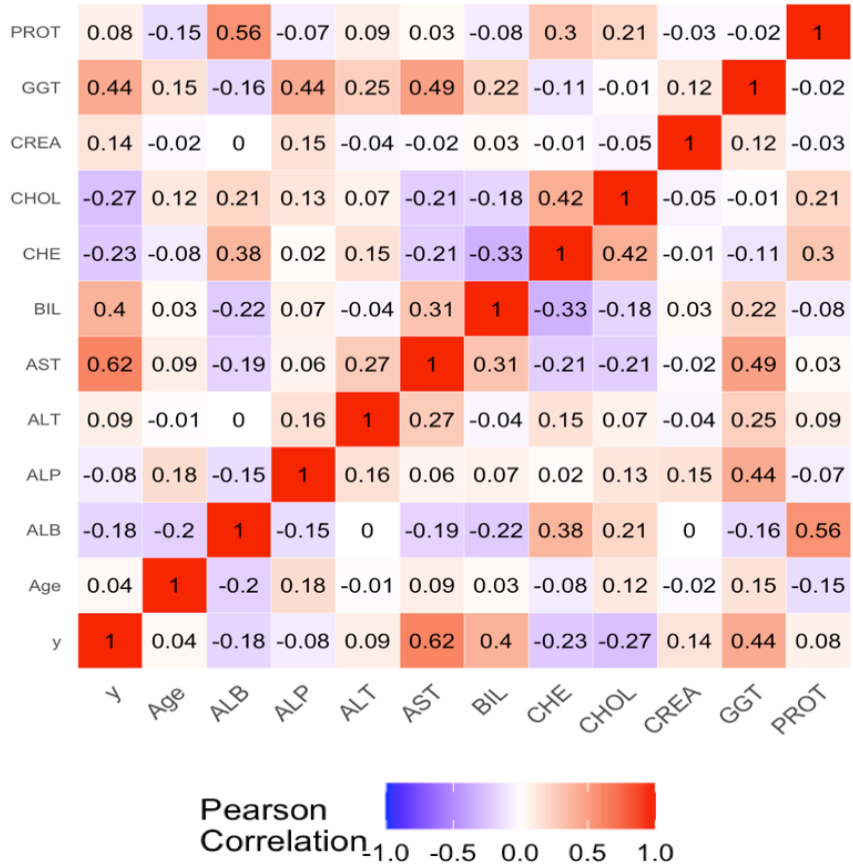

**Figure 6: Correlation matrix of all covariates and response variables**

Target variable $y$ has moderately positive relationships with AST, BIL and GGT. However, y has fairly weak relationships with ALB and CHE. However, CHOL, PROT and Age variables do not impact much.

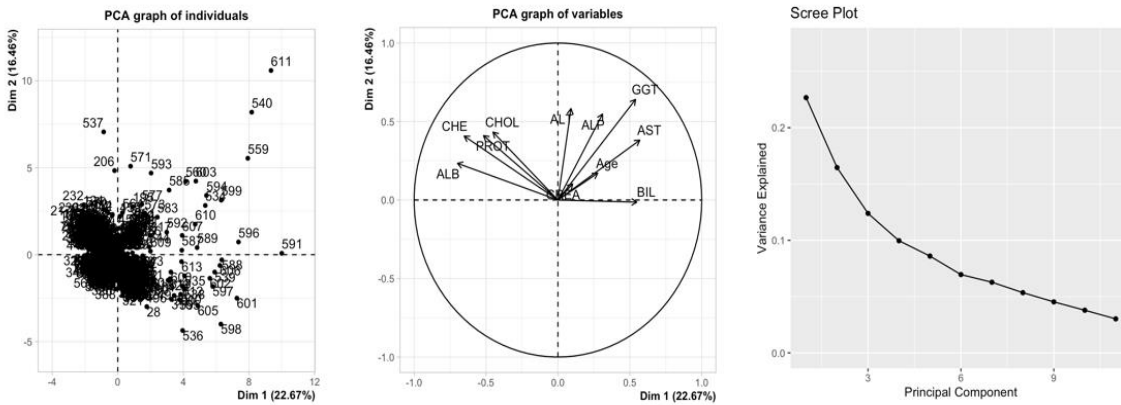

**Figure 7: Dimension reduction using PCA.**

For data reduction, a nice result is obtained for our study. It reduces the dimensionality by projecting each data point into the first few principal components. **In Figure 7**, a scree plot (right) has been shown to decide the number of predictors. Results in **Figure 7** and principal components— AST, ALT, ALP, BIL, CHE and GGT show almost 90% variability.

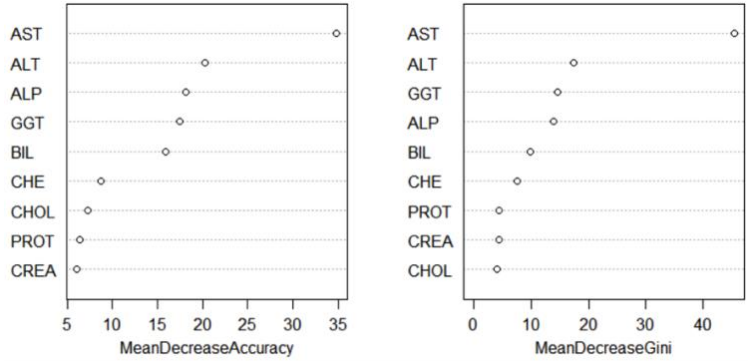

**Figure 8: Variable importance ranking in RF**

Moreover, the variable importance ranking is obtained for RF and it is measured using mean decrease accuracy and mean decrease gini as parameters. Hence, from **Figure 8**; AST, ALT, ALP, and GGT are the most important variables observed in the dataset. After comparing with Figure 05, and results from PCA; AST, ALT, ALP, and GGT are used to train classification models.

**Table 1: Model Evaluation**

| Model | ANN | SVM | RF |
|---|---|---|---|
| **Sensitivity** | 0.9960 | 1.00 | 0.9859 |
| **Specificity** | 0.8116 | 0.8551 | 0.8261 |
| **Accuracy** | 0.9734 | 0.9823 | 0.9663 |
| $F_1$ | 0.9851 | 0.9891 | 0.9863 |

According to **Table 1**, SVM shows the highest sensitivity value 1, where highest accuracy value was reported for ANN. Results also indicate the Sensitivity which evaluates the described model's ability to predict true positives of each available category, and Specificity evaluates model's ability to predict true negatives of each available category. In case of RF,

Table 2: Confusion matrix with Class Prediction

|  |  | Actual Class | | | Actual Class | | | Actual Class | |
|---|---|---|---|---|---|---|---|---|---|
|  | ANN | 0 | 1 | SVM | 0 | 1 | RF | 0 | 1 |
| **Predicted Class** | 0 | 493 | 13 | 0 | 495 | 10 | 0 | 488 | 12 |
|  | 1 | 2 | 56 | 1 | 0 | 59 | 1 | 7 | 57 |

**Table 2** presents; 7 samples lied in FP and 12 samples lied in false negative. An interesting result is observed for SVM; there are no FP samples in our data set. The maximum specificity value was found for SVM between all models (**see Table 1**). ANN takes more time than SVM and RF; between all these three studied models SVM is faster.

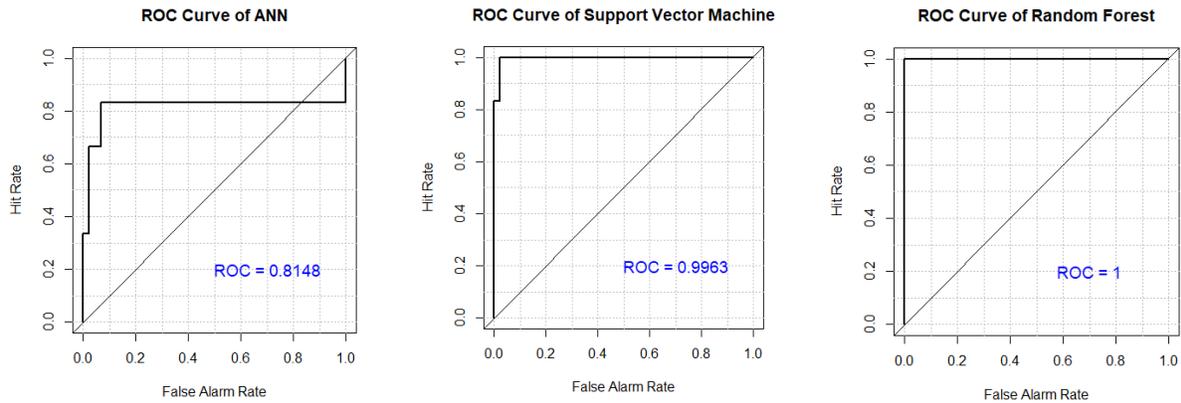

Figure 9: ROC Curves for ANN, SVM and Random Forest classifiers

According to our results, ML can support health sectors to achieve efficient and more effective results for identifying groups or levels. Higher accuracy means better diagnosis and management.

Confusion Matrix used in this research can be used to help health workers to make decisions. Moreover, ML methods are data driven, so it directly uses health records of particular patients. Thus, this process is more reliable. The **AUC-ROC** curve of the classification validates our applied techniques and a good accuracy level is found. Moreover, 0-1 loss function supports our results where SVM shows lowest 1.77% expected loss, which is very low.

## 3. CONCLUSIONS

This study showed some machine learning approaches with PCA which outperformed the performance of classification. Between three techniques, our proposed models such as SVM and RF performed better than deep learning method ANN. Although, accuracy levels for all models were excellent. The machine learning algorithms presented here can support but it is not alternative to the medical expert when designing decisions from ML classifiers for diagnostic pathways. Our methods can save time and costs for the betterment of people. Instead of using 10-fold cross validation for splitting the dataset into training and test datasets, one may use random data splitting or stratified sampling of 3:1 for obtaining training and test datasets. In that way one can compare the performance of the models for determining healthy blood donors and separate liver disease patients. We had to apply binary classification by combining all the unhealthy blood donors due to the shortage of the observations for Hepatitis (24), Fibrosis (21), and Cirrhosis (30). Hence, using the multinomial classification we could compare the obtained results. Another future research direction of this study is to check other machine learning techniques for comparing the obtained results since we obtained excellent accuracy rates to figure out the healthy blood donor. This study does not claim that traditional medical diagnosis can be replaced by ML methods, but these can be used to obtain better diagnosis.

**ACKNOWLEDGEMENT**

All authors contributed equally to this work. We are very grateful to the unknown reviewers for their valuable suggestions. This project is not funded by any company or institution.